\documentclass[letterpaper, 10 pt, conference]{ieeeconf}  

\IEEEoverridecommandlockouts                              
\usepackage{comment}
\usepackage{mathbbol}
\usepackage{tabularx}
\usepackage{algorithm}
\usepackage{xcolor, soul}
\sethlcolor{green}

\let\labelindent\relax

\usepackage{etoolbox}
\makeatletter
\patchcmd{\@makecaption}
  {\scshape}
  {}
  {}
  {}
\makeatother
\usepackage[caption=false]{subfig}

\usepackage{algorithmic}
\usepackage{hyperref}
\usepackage{euscript}[mathcal]
\usepackage[inline]{enumitem}
\usepackage{multirow}
\usepackage{amsmath}
\usepackage{esvect}
\usepackage{color}

\usepackage{url}
\usepackage{xcolor}
\hypersetup{
    colorlinks=true,
    linkcolor=blue,
    filecolor=magenta,      
    urlcolor=cyan,
}

\definecolor{newcolor}{rgb}{0.858, 0.188, 0.478}
\usepackage[english]{babel}
\usepackage[caption=false]{subfig}
\usepackage{multirow}
\usepackage{wrapfig}
\usepackage{booktabs}
\usepackage{array}

\usepackage{stfloats}
\usepackage[utf8]{inputenc}
\usepackage[T1]{fontenc}
\usepackage{amsmath}

\usepackage{cite}
\input epsf
\usepackage{graphicx}
\usepackage{adjustbox}
\graphicspath{{./images/}}
\usepackage{amssymb}
\usepackage{longtable}

\usepackage{caption} 

\usepackage{hyperref}
\usepackage{cleveref}

\makeatletter 
\@namedef{ver@float.sty}{3000/12/31}
\makeatother

\usepackage{mathptmx} 
\usepackage{amssymb}  

\title{\LARGE \bf
A Robust Completed Local Binary Pattern (RCLBP) for Surface Defect Detection}

\author{Nana Kankam Gyimah$^{*}$, Abenezer Girma$^{*}$, Mahmoud Nabil Mahmoud$^{*}$, \\
Shamila Nateghi$^{*}$,
Abdollah Homaifar$^{*}$,
Daniel Opoku$^{+}$\\
$^{*}$ North Carolina A\&T State University, Greensboro, North Carolina, US, 27411\\ 
$^{+}$ Kwame Nkrumah University of Science $\&$ Technology, Kumasi, Ghana\\

\thanks{Corresponding author: 
  Tel.: +1-336-285-3271;   fax: +1-336-334-7716;  \textit{e-mail: homaifar@ncat.edu} (Abdollah Homaifar)}
  }

\begin{document}

\maketitle
\thispagestyle{empty}
 \pagestyle{empty}

\begin{abstract}
In this paper, we present a Robust Completed Local Binary Pattern (RCLBP) framework for a surface defect detection task. Our approach uses a combination of Non-Local (NL) means filter with wavelet thresholding and Completed Local Binary Pattern (CLBP) to extract robust features which are fed into classifiers for surface defects detection. This paper combines three components: A denoising technique based on Non-Local (NL) means filter with wavelet thresholding is established to denoise the noisy image while preserving the textures and edges. Second, discriminative features are extracted using the CLBP technique. Finally, the discriminative features are fed into the classifiers to build the detection model and evaluate the performance of the proposed framework. The performance of the defect detection models are evaluated using a real-world steel surface defect database from Northeastern University (NEU). Experimental results demonstrate that the proposed approach RCLBP is noise robust and can be applied for surface defect detection under varying conditions of intra-class and inter-class changes and with illumination changes.

Index Terms - Surface defect, Intra-class defect differences, Inter-class defect similarities,  Non-local means filter with wavelet thresholding, Completed Local Binary Pattern (CLBP)
\end{abstract}

\section{Introduction}
Metal planar materials (steel, aluminum, copper plates, and strips) are widely used in aerospace, automobile manufacturing, bridge construction, and other pillar industries. These industries have made
immense contributions to modern social development and the improvement of life \cite{fang2020research}. The occurrence of surface defects (corrosion, cracks, scratches or dents) on these metal planar materials during the manufacturing process and use of these industrial products can cause huge economic losses when these defects are left undetected \cite{sun2018research}. 

These surface defects have traditionally  been  detected by professionally trained human inspectors \cite{zhou2019automatic} who judge the quality of the defect by observing and inspecting the differences of the surface appearance of the product with the naked eyes. This method however does not meet the needs of modern industrial production since the criteria for human vision is not quantified but rather dependent on the subjective evaluation of the human inspectors. The quality of this inspection is not fully guaranteed and less efficient since human inspectors can experience ocular fatigue due to the high intensive and repetitive nature of work which leads to less reliable defect detection. 

Automatic Visual Inspection System (AVIS), however, has gained greater acceptance with the development of automated Computer Vision (CV) based inspection \cite{mera2016automatic}. CV based inspection methods have significantly increased the defect detection speed, reduced labor costs, and improved the quality of inspection through effective damage localization \cite{malamas2003survey}. CV based inspection have improved the productivity  \cite{malamas2003survey} of defect detection as they provide a competitive advantage to the traditional human inspection process for detecting surface defects.

\begin{figure*}
\includegraphics[scale=0.4]{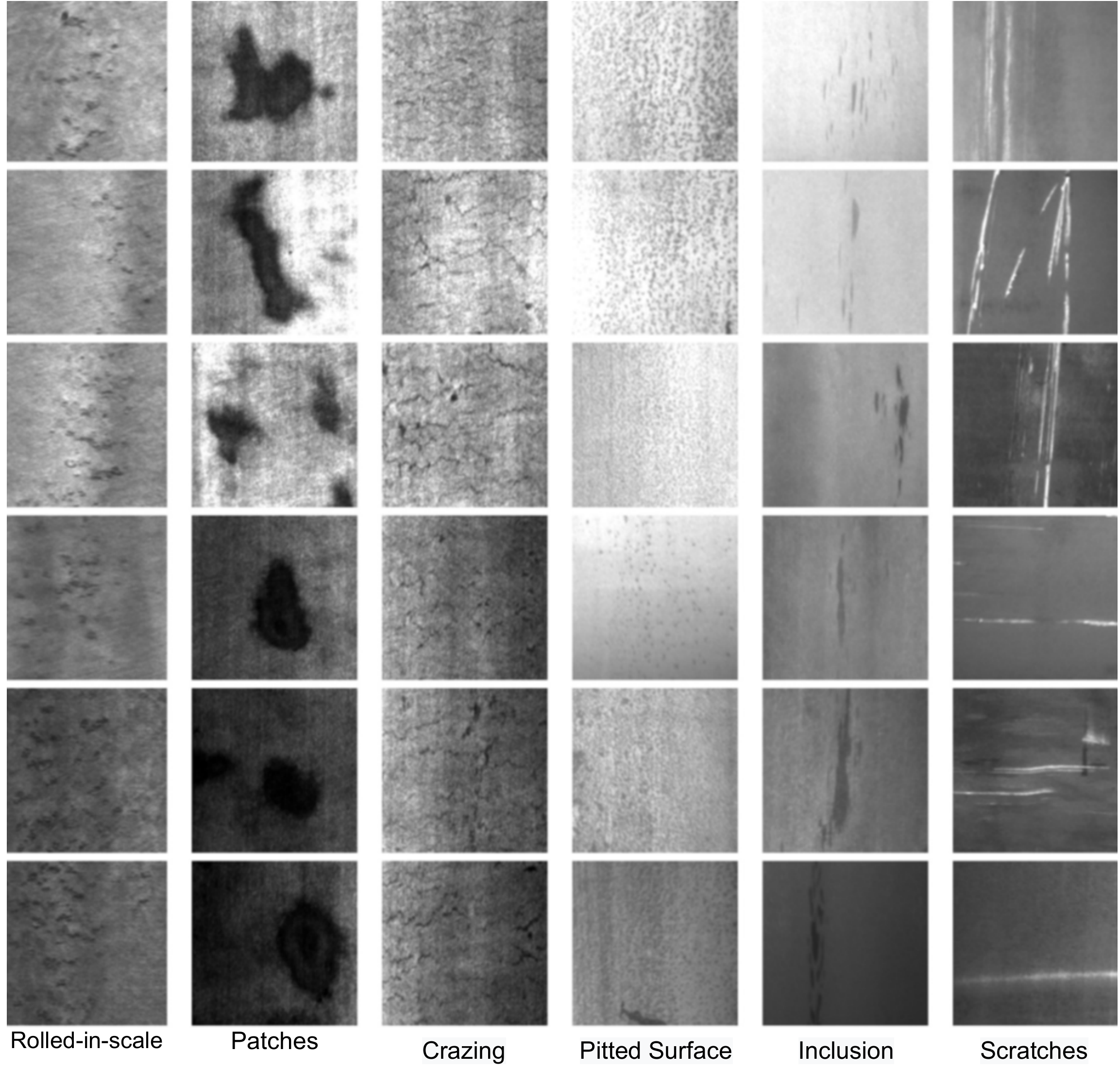}
\centering
\caption{Samples of the defect classes of the Northeastern University (NEU) surface defect dataset\cite{song2013noise}. The columns define the defect class with the rows showing the sample images that belongs to each defect class.}
\label{fig:imageee}
\end{figure*}

The CV-based surface defect detection however has three main challenges which tends to diminish the recognition performance of defect detection approaches and affects the stability of the extracted defect features as outlined below:
\begin{enumerate}
    \item Distinguishing between the inter-class defect similarities and intra-class defect differences. Inter-class defects similarities refer to different defect classes that appear nearly identical in their appearance as shown between the defect classes of rolled-in scale and crazing in Fig. \ref{fig:imageee}. Intra-class defect differences refer to the defect class whose images are random and highly diverse in their appearance as in the defect classes of scratches and patches in Fig. \ref{fig:imageee}.
    \item Background interference due to the influence of different illumination and material changes tend to reduce the recognition performance of the defect detection approaches.
    \item Most of the employed features, the Local Binary Pattern (LBP) \cite{ojala2002multiresolution,guo2010completed} and Multi-Scale Geometric Analysis (MGA) features \cite{xu2013application} are vulnerable to noise which reduces the recognition performance of these detection approaches.
\end{enumerate}

In \cite{guo2010completed}, the Completed Local Binary Pattern (CLBP) was proposed to characterize discriminant information that describes defect classes. CLBP is able to distinguish between the interclass defects, the intraclass defects and also address the problem of background interference due to its grayscale and rotation invariance. However, it is sensitive to noise and has limited ability to represent miscellaneous textures. In \cite{xu2013application}, multi-scale geometric analysis (MGA) features was proposed which was effective for classifying surface defects. However, it is highly sensitive to noise.
In \cite{lin2019automated}, LEDNet based on Convolutional Neural Network (CNN) was proposed to detect and classify defects on LED chips. This method achieved high defect detection accuracy but sensitive to noise.

To overcome the challenges in existing defect detection approaches, this paper presents a robust defect detection framework which aims to address the three challenges that affects the stability of the extracted features and the recognition performance of the defect detection approaches. The contributions of this framework are as follows:
 \begin{enumerate}
    \item The denoising technique based on the combination  of Non-Linear (NL)-means filter with wavelet thresholding is proposed as a pre-processing step to remove the noise while retaining the input image's textures and edges as much as possible to address the vulnerability of these features to noise.
    \item CLBP is explored to address the intraclass defect differences, interclass defect similarities, and background interference challenges because of its grayscale and rotation invariance.
    \item Extensive experiments on the NEU-DET dataset demonstrate the effectiveness of the proposed method in the presence of additive Gaussian noise compared to Local Binary Pattern (LBP), and Completed Local Binary Pattern (CLBP).
\end{enumerate}

The rest of this paper is organized as follows: Section $\text{II}$ presents related works on defect detection. Section $\text{III}$ describes the proposed methodology. Section $\text{IV}$ describes the dataset, experiment setup, hyper-parameter settings and the evaluation Metrics. Section $\text{V}$ presents the experimental results and discussion. Finally, the conclusion and future work are provided in Section $\text{VI}$.

\section{Related Works}
This section describes relevant literature that focuses on defect detection methods. Automated Surface Inspection (ASI) task has been modelled as a texture analysis problem where the surface defects are generally described as local anomalies in homogeneous textures. In general, surface defect detection methods based on texture analysis are decomposed into statistical-based methods \cite{shi2016improved}, spectrum-based methods \cite{sharifzadeh2008detection}, model-based methods \cite{luo2018generalized}, and emerging machine learning-based methods \cite{duan2016deep}.

\textbf{Statistical methods}: For this approach, the regular and periodic distribution of pixel strength is investigated by measuring the statistical characteristics of pixel spatial distribution to detect the defects on the metal planar materials surface. These methods are based on edge detection \cite{shi2016improved}, hough transform \cite{sharifzadeh2008detection}, gray-level statistics \cite{ma2017surface}, local binary pattern \cite{song2013noise} and Generalized Completed LBP (GCLBP) \cite{luo2018generalized}. In \cite{shi2016improved}, edge detection based on eight directional Sobel operators was utilized to detect Backfin defects. This was robust to noise and protected the edge shape but was only suitable to low resolution images. In \cite{sharifzadeh2008detection}, Hough Transform (HT) was utilized to detect holes, scratches, coil breakage, and corrosion on a cold-rolled steel strip. HT has strong anti-interference ability which suppresses the influence of noise and incomplete edges. However, the uncertainty of the defect's shape of most metal planar materials often leads to unsatisfactory detection. In \cite{ma2017surface}, multi-directional gray fluctuation was used to characterize multi-type defects. This method was suitable for low-resolution images but cannot automatically select the threshold to extract the defect features. In \cite{luo2018generalized}, the Generalized Completed LBP (GCLBP) was utilized to explore the non-uniform pattern hidden in the uniform pattern to detect the multi-class defect types. This method has strong anti-interference ability but it cannot suppress noise and is also unable to simultaneously adapt to scale variation at the same time. 

\textbf{Structural methods}: These approaches model the texture elements and spatial arrangements that characterize the defects. These methods are based on Fourier Transform \cite{ai2013surface}, Gabor filter \cite{choi2011pinhole}, and Wavelet Transform \cite{wu2008application}. In \cite{ai2013surface}, a combination of fourier transform and curvelet transform was proposed to detect longitudinal cracks. This method is invariant to translation, expansion, and rotation. However, the background and defect information in frequency domain can easily be mixed to cause interference. In \cite{choi2011pinhole}, traditional Gabor filter was utilized to detect periodic defects. It is suitable for high-dimensional feature space, but quite difficult to determine the optimal filtering parameters which provides no rotation invariance. In \cite{wu2008application}, the undecimated wavelet transform was used to detect horizontal scratch defects. It is suitable for multi-scale image analysis and can compress images effectively. However, it is difficult to select the proper wavelet base to extract the defect features.

\textbf{Model-based methods}: These approaches projects the original texture distribution of the image block to the low-dimensional distribution through the special structure model enhanced by parameter learning to detect various defects. These approaches are based on Markov random field \cite{luo2020automated}, visual saliency model  \cite{zhou2018double} and fractal dimension model \cite{yazdchi2009steel}. In \cite{luo2020automated}, the hidden Markov tree model was utilized to detect the multi-type defects. This method is able to reflect the underlying structure of the image but is not suitable for global texture analysis and small size defects. In \cite{zhou2018double}, the double low-rank and sparse decomposition was utilized to detect multi-type defects. It is robust to noise and uneven illuminations but limited to gradient strength or low contrast defects. In \cite{yazdchi2009steel}, multi-fractal decomposition was used to detect multi-type defects. The global information can be represented by the local features but is only applicable to images with adaptability. 

\textbf{Emerging machine learning-based methods}: These methods are data driven approaches that have been categorized into supervised learning \cite{kang2005surface}, unsupervised learning \cite{bulnes2012vision} and semi-supervised learning \cite{di2019surface}. These approaches are effective however, they require a relatively high number of training data and are computationally expensive.

The aforementioned methods addresses either one or two of the challenges to detect the surface defects. To solve these challenges, this paper presents a noise robust defect detection framework which is rotation and grayscale invariant for surface defect detection. The proposed framework differs from LBP \cite{ojala2002multiresolution} and CLBP \cite{guo2010completed} through the use of the denoising technique.

\section{Proposed Methodology}
\begin{figure*}
\includegraphics[scale=0.3]{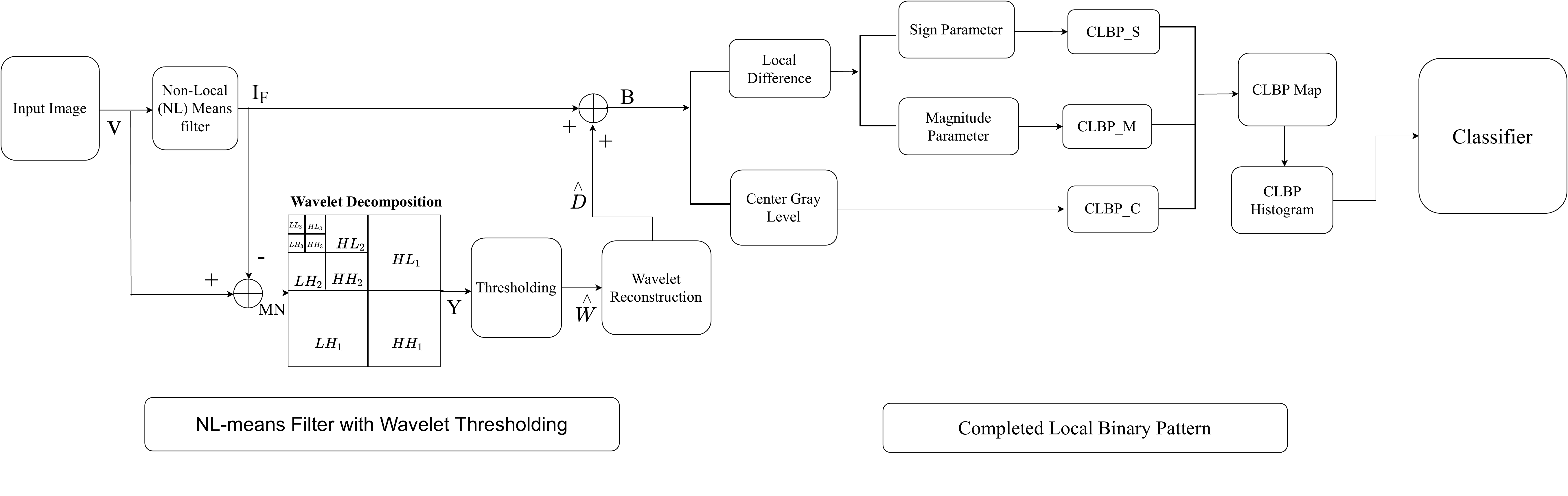}
\caption{The Robust Completed Local Binary Pattern (RCLBP) Framework.}
\label{fig:imagee77}
\end{figure*}

The proposed framework consists of two distinct stages as shown in Fig \ref{fig:imagee77}. The first stage is the denoising stage which is a combination of Non-Local (NL) means filter with wavelet thresholding \cite{kumar2013image} to remove noise from the images. The second stage is the feature extraction process where we extract the features which characterize the defect classes using CLBP \cite{guo2010completed}. 

\subsection{Denoising}
The goal of image denoising is to remove the noise while retaining the image textures and edges as much as possible. From Fig. \ref{fig:imagee77}, the input image $V$ is first denoised by the NL-means filter \cite{buades2005review}. The NL-means filter is effective in removing the noise at high Signal-to-Noise Ratio (SNR) (less noise) while retaining the image's textures and edges. However as the noise increases (low SNR), its performance deteriorates because it wrongly considers the image's textures and edges as noise and removes them from the features \cite{kumar2013image}. To preserve these important information while removing the noise, the difference between the input image ($V$) and the NL-means filtered image ($I_{F}$) is computed similar to the Method Noise ($MN$) \cite{kumar2013image} as $MN$ contains the input image's textures and edges. The $MN$ however contains noise. To estimate the input image's textures and edges $\overset{\wedge}D$, the method noise undergoes wavelet transformation to obtain subbands $Y$ which are thresholded with BayesShrink filter to estimate the true wavelet subbands $\overset{\wedge}W$ without noise. These denoised wavelet subbands are subsequently reconstructed to estimate the image's textures and edges $\overset{\wedge}{D}$. The estimated image's textures and edges $\overset{\wedge}{D}$ is finally fused with the NL-means filtered image $I_{F}$ to generate $B$ which is the denoised image with restored textures and edges. The denoising framework is as discussed in Section \ref{section:my}. 

\subsubsection {\textbf{Denoising Framework}}
\label{section:my}
Given a discrete noisy image $V=\{ V(i)|i \in I\}$ on a discrete grid $I \in \mathbb{R^2}$, the estimated restored intensity $NL(i)$ \cite{buades2005review} for a pixel ($i$), is computed as a weighted average of the pixel intensities $v(j)$ of the image $V$ as defined in Eq. \eqref{eqn:ssomelabel}.
\begin{align}
NL(i) = \sum_{j\in I} w(i,j)v(j)
\label{eqn:ssomelabel}
\end{align}
where ${w(i,j)}$ is the weight assigned to $v(j)$ for restoring the pixel $i$. The number of pixels $(j)$ in the weighted average is restricted to a neighborhood search window $S_{i}$ centered at the pixel $i$. 

To compute the similarity between image pixels, a neighborhood patch $N_{k}$ of fixed square size centered at a pixel $k$ and within the search window $S_{i}$. The weights ${w(i,j)}$ evaluates the similarity between the intensities of the local neighborhoods  
centered on pixels $i$ and $j$.

To compute the similarity of the intensity gray level vectors $v(N_{i})$ and $v(N_{j})$, we compute a Gaussian weighted Euclidean distance, $\|v(N_{i})-v(N_{j})\|_{2,\sigma}^{2}$, where $\sigma>0$ is the standard deviation of the Gaussian kernel. The Euclidean distance is the traditional $L_{2}$-norm convolved with a Gaussian kernel of standard deviation $\sigma$ as it  preserves the order of similarity between pixels. The weights $w(i,j)$ is computed as in Eq. \ref{eqn:somelabelll}
\begin{align}
w(i,j)=\frac{1}{Z(i)} e^-\frac {\|v(N_{i})-v(N_{j})\|_{2,\sigma}^{2}}{h^{2}}
\label{eqn:somelabelll}
\end{align}

where Z(i) is the normalizing factor as in Eq. \ref{eqn:somelabl}
\begin{align}
Z(i)=\sum_{j} e^-\frac {\|v(N_{i})-v(N_{j})\|_{2,\sigma}^{2}}{h^{2}}
\label{eqn:somelabl}
\end{align}
which ensures that $\sum_{j} w(i,j)=1$ and \emph{h} is the smoothing kernel width which controls the decay of the exponential function.

To estimate $I_{F}$, all the noisy pixels in $V$ are restored with the estimated intensity NL(i) as defined in Eq. \ref{eqn:NL-filtered-image}.
\begin{align}
I_{F} = \{ NL(i)|i \in I\}
\label{eqn:NL-filtered-image}
\end{align}
The application of the NL-means filter on the noisy image removes the noise and cleans the edges without losing too many textures and edges.

At low SNR, the NL-means filtered image $I_{F}$ considers the input image's textures and edges as noise and removes them from the features. To restore $I_{F}$ with its lost textures and edges, the method noise ($MN$) is computed as the difference between the noisy image ($V$) and the NL-means filtered image ($I_{F}$) which is defined in Eq. \ref{eq:method_noise}.
\begin{align}
MN = V-I_{F}
\label{eq:method_noise}
\end{align}
where $v$ is the original image, $I_{F}$ is the $NL$-means filtered image of the input image $V$ and $MN$ is the Method Noise. 

The method noise undergoes a wavelet transformation where MN is decomposed into k number of sub-bands; $HH_{k}$, $HL_{k}$, $LH_{k}$, $LL_{k}$ where $k$=\{1, 2, $\cdots$, $J$\} is the scale, with J being the largest scale of the decomposition. The sub-bands $HH_{k}$, $HL_{k}$, and $LH_{k}$ represents the textures and edges sub-band of MN whiles $LL_{k}$ represents the approximate sub-band of MN. These noisy wavelet sub-bands referred to as Y in Fig. \ref{fig:imagee77} is a combination of the sub-bands textures and some Gaussian noise N as defined in Eq. \ref{eq:wavelet method_noise}.

\begin{align}
Y = \overset{\wedge}W+N
\label{eq:wavelet method_noise}
\end{align}

To estimate $\overset{\wedge}W$, the denoised subbands textures from $Y$, the subbands of Y are thresholded with the adaptive BayesShrink threshold \cite{chang2000adaptive} to remove the noise N by minimizing the Mean Squared Error (MSE) between $\overset{\wedge}W$ and Y. The BayesShrink threshold ($T$) which is adaptive to each wavelet sub-band yields a data driven estimate of the threshold $T$ through soft thresholding as defined in Eq. \ref{eq:BayesShrink} to remove the noise.
\begin{align}
W_{soft}= 
    \begin{cases}
      sgn(Y)(|Y|-T), |Y|> \text{$T$}\\
      0, |Y|\leq \text{$T$}\\
    \end{cases} where \quad
    T = \frac{\sigma^{2}}{\sigma_{w}}
    \label{eq:BayesShrink}
\end{align}
where sgn depicts the signum function, $W_{soft}$ is the wavelet coefficient after the shrinkage of the soft threshold, and $\sigma^{2}$ is the noise variance estimated from subband $HH_{1}$ by a robust median estimator \cite{donoho1994ideal} as defined in Eq. \ref{eq:BayesssShrink}.
\begin{align}
{\sigma^{2}}=(\frac{Median(|Y_{i,j}|)}{0.6745})^{2},  Y_{i,j} \in {HH_{1}}
\label{eq:BayesssShrink}
\end{align}
For the wavelet model given in Eq. \ref{eq:wavelet method_noise}, then  
\begin{align}
{\sigma_{Y}^{2}}=\sigma_{W}^{2}+\sigma^{2}
\label{eq:BayyessShrink}
\end{align}
where $\sigma_{Y}^{2}$ is the variance of Y, $\sigma_{W}^{2}$ is the variance of $\overset{\wedge}W$, and $\sigma^{2}$ is the variance of the Gaussian noise N. Since Y is modelled as zero mean, its variance $\sigma_{Y}^{2}$ can be found empirically as defined in Eq. \ref{eq:variance_Y}. 
\begin{align}
{\sigma_{y}^{2}} = \frac{1}{MN} \sum Y^{2}
\label{eq:variance_Y}
\end{align}
where $M$ and $N$ is the size of the subband of Y under consideration.
The variance, $\sigma_{w}^{2}$ of the wavelet sub-bands $\overset{\wedge}W$ is finally computed as defined in Eq. \ref{eq:variance}.
\begin{align}
{\sigma_{w}^{2}}=max({\sigma_{y}^{2}}-{\sigma^{2}}, 0)
\label{eq:variance}
\end{align}

The denoised subbands of $\overset {\wedge}W$ are wavelet reconstructed to give an estimate of the detail image $\overset {\wedge}D$ containing the lost textures and edges.

To estimate the denoised image with the restored textures and edges $B$, the detailed image $\overset {\wedge}D$ from the wavelet thresholding is summed up with the NL-means filtered image $I_{F}$ as defined in Eq. \ref{eq:NLFMT} as it restores $I_{F}$ with the lost textures and edges.
\begin{align}
B=I_{F}+\overset {\wedge}D
\label{eq:NLFMT}
\end{align}

\subsection{Feature Extraction}
\begin{figure}
\includegraphics[scale=0.7]{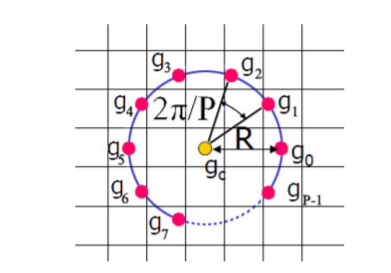}
\caption{ Central pixel ($g_{c}$) with its  circularly and evenly spaced neighbors P with radius R of the neighborhood \cite{guo2010completed}.}
\label{fig:imagee3}
\end{figure}

Completed Local Binary Pattern (CLBP) \cite{guo2010completed} is used to extract the discriminative features that characterize the features of the denoised image B as defined in Eq. \ref{eq:NLFMT}. It represents the local region of the denoised image as shown in Fig. \ref{fig:imagee3} by the local difference sign-magnitude transform (LDSMT) and its center pixel as shown in Fig. \ref{fig:imagee77}. 

Given a central pixel $g_{c}$ with $P$ circularly and evenly spaced neighbors $g_{p}$, $p=0, 1, \cdots, P-1$ in Fig. \ref{fig:imagee3}, the local difference between $g_{c}$ and $g_{p}$ is computed as $d_{p}=g_{p}-g_{c}$. The local difference characterizes the local structure at $g_{c}$ and can be decomposed into the sign ($s$) and magnitude ($m$) components as defined in Eq. \ref{eq:LSMD}.
\begin{align}
d_{p}=s*m, \quad \textrm{and} \quad
    \begin{cases}
      s= \text{$sign(d_{p})$}\\
      m= \text{$|d_{p}|$}\\
    \end{cases}
\label{eq:LSMD}
\end{align}

To encode the sign component ($s$) of the local difference ($d_{p}$), $CLBP\_S$ is proposed to extract the sign information by comparing the pixel with its neighboring pixels as defined in Eq. \ref{eq:sign}.
\begin{align}
CLBP\_S=\sum_{p=0}^{P-1} s(g_{p}-g_{c})2^{p}, \quad \quad
s(x) =
    \begin{cases}
      1, & \text{$x\geq0$}\\
      0, & \text{$x<0$}\\
    \end{cases}
\label{eq:sign}
\end{align}
where $g_{c}$ is the gray value of the central pixel, $g_{p}$ is the pixel value of its neighbors, $P$ is the total number of neighboring pixels and $R$ is the radius of the neighborhood.

To extract the local variance information contributed by the magnitude ($m$) component of the local difference ($d_{p}$), $CLBP\_M$ is proposed to code the magnitude component of the local information as defined in Eq. \ref{eq:magnitude}.
\begin{align}
CLBP\_M=\sum_{p=0}^{P-1} t(m_{p},c)2^{p}, \quad \quad 
t(x,c) =
    \begin{cases}
      1, & \text{$x\geq c$}\\
      0, & \text{$x<c$}\\
    \end{cases}
\label{eq:magnitude}
\end{align}
Where, $c$ is an adaptive threshold which is set to be the mean value of $m$ from the whole denoised image.

The center pixel ($c$) which expresses the image local gray level also has discriminant information. To make it consistent with $CLBP\_S$ and $CLBP\_M$, $CLBP\_C$ is defined in Eq. \ref{eq:Center} to encode the center pixel (C) information. 
\begin{align}
CLBP\_C=t(g_{c},c_{I}), \quad \quad 
t(x,c_{I}) =
    \begin{cases}
      1, & \text{$x\geq c_{I}$}\\
      0, & \text{$x<c_{I}$}\\
    \end{cases}
\label{eq:Center}
\end{align}
where the threshold $c_{I}$ is set as the average gray level of the denoised image.

The three operator ma[s, $CLBP\_S$, $CLBP\_M$ and $CLBP\_C$  are combined jointly to generate the CLBP histogram as the input to the classifiers as shown in Fig. \ref{fig:imagee77}.

\section{Dataset, Experiment Setup, Hyper-parameter tuning and Evaluation Metrics}
This section describes the dataset and the Gaussian noise addition process. It also describes the experiment setup, the hyper-parameter tuning and the performance evaluation metrics of the defect detection model.

\subsection{Dataset}
\label{subsection:Dataset}
The dataset used for the experiments is the NorthEastern University (NEU) surface defect dataset \cite{song2013noise}. It comprises of six defect classes: rolled-in scale ($Rs$), patches ($Pa$), crazing ($Cr$), pitted surface ($Ps$), inclusion ($In$) and scratches ($Sc$). The collected defects are on the surface of the hot-rolled steel strip. The dataset consists of 1800 gray-scale images: 300 labelled samples for each defect class. Fig. \ref{fig:imageee} shows some sample images of the six defect classes.

\begin{figure}
\includegraphics[scale=0.5]{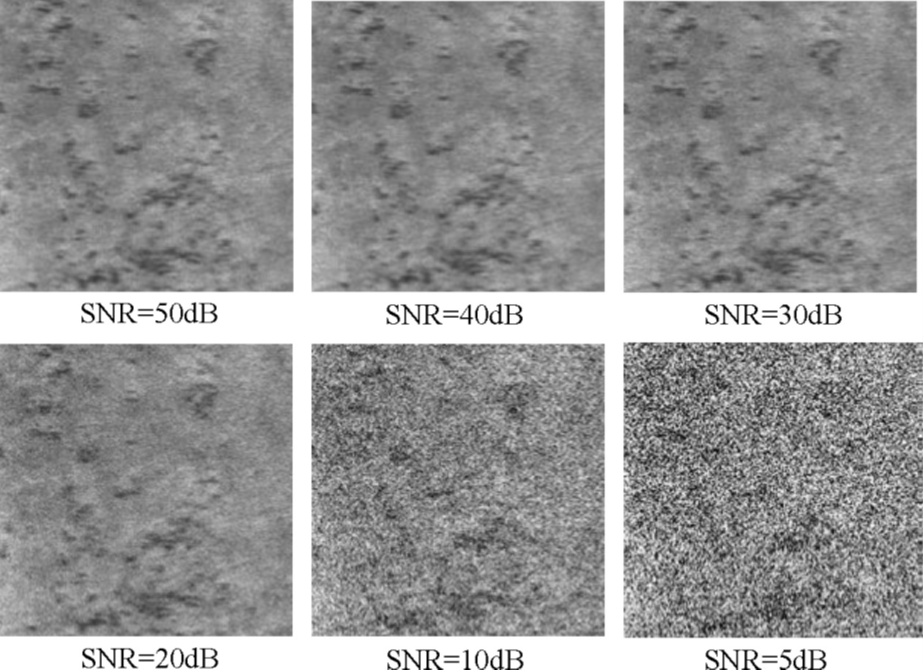}
\caption{Surface defect images with Gaussian noise of different SNR \cite{song2013noise}.}
\label{fig:imagee-noise}
\end{figure}

To evaluate the robustness of the proposed approach against noise, noisy defect images are generated with the addition of Gaussian noise of different SNR. Fig. \ref{fig:imagee-noise} shows some defect images with some added Gaussian noise of different SNR. 
It can be observed from Fig. \ref{fig:imagee-noise} that, the texture appearance of the defect samples with SNR of 50dB and 40dB are rarely affected by the noise. At SNR of 30dB, there is moderate change in the textural appearance of the defect samples. However, there is a significant change in the textural appearance of the defect samples when the SNR of the Gaussian noise is at most $20$dB. 

\subsection{Experiment Setup}
The dataset used for the experiments is described in section \ref{subsection:Dataset}. 
This dataset is split into $80\%$ training and $20\%$ testing set.
To evaluate the robustness of the proposed method against additive Gaussian noise, the proposed model is trained with the noise free training dataset, and tested on the Gaussian noise induced test dataset. Different SNR level of noise from $50dB$ to $20dB$ is introduced on the test set. To evaluate the performance of the proposed framework, K-Nearest Neighbor (KNN), Support Vector Machine (SVM), Decision Tree (DT), Gaussian Naive Bayes (GNB) and Random Forest (RF) classifiers \cite{itoo2020comparison} are used to evaluate the performance of the defect detection model.

\subsection{Hyper-parameter Tuning}
The hyper-parameters of the classifiers used for the comparison of the defect detection models are determined through a grid search and a $K$ fold cross validation with a $K$ value of 10. The K-fold cross validation is implemented on the $80\%$ training dataset.The hyper-parameters of the various classifiers are as summarized in Table \ref{table:validation accuracy}.

\begin{table}[h!]
\caption{Classifiers Hyper-parameter Settings.}  
\centering 
\footnotesize
{\begin{tabular}{c c c} 
\hline
Classifier &Hyper-parameter &Values\\ [0.3ex]
\hline
 &$n_{-}neighbors$ &$9$\\
KNN &p &$1$ \\
 &$leaf_{-}size$ &$10$\\
 &weights &distance\\
\hline
&C &0.1\\
SVM &Gamma &1 \\
  &Kernel &Linear\\
\hline
 &Criterion &Gini \\
 &$max_{-}depth$ &$9$ \\
DT &$max_{-}features$ &sqrt\\
 &$min_{-}samples_{-}leaf$ &$1$ \\
 &$min_{}-samples_{-}split$ &$2$\\
\hline
GNB &$var_{-}smoothing$ &$3.51e^{-08}$\\
\hline
 &Criterion &Entropy \\
 &$max_{-}depth$ &$9$ \\
RF &$max_{-}features$ &sqrt\\
 &$min_{-}samples_{-}leaf$ &$7$ \\
 &$min_{}-samples_{-}split$ &$4$\\
 &$n_{}-estimators$ &$10$\\
\hline 
\end{tabular}}
\label{table:validation accuracy}
\end{table}
\subsection{Evaluation Metrics}
\label{table:evaluation metrics}
The weighted average values of Precision, Recall and F1-score measures as defined in Eq. \ref{eq:precision} through Eq. \ref{eq:F1-score} are adopted as the metrics to evaluate the performance of the detection models\cite{zhang2021image}. These metrics are defined by means of the true positive ($tp_{i}$), true negative ($tn_{i}$), false positive ($fp_{i}$) and false negative ($fn_{i}$) of each defect class $C_{i}$ with $i=1, \cdots, m$ where $m$ is the total number of defect classes in the dataset. $|Y_{i}|$ is the  total number of samples assigned to each defect class.
\begin{align}
Weighted\ Average\ Precision =\frac{\sum_{i=1}^{m}|Y_{i}| \frac{tp_{i}}{tp_{i} + fp_{i}}}{\sum_{i}^{m}|Y_{i}|} 
\label{eq:precision}
\end{align}

\begin{align}
Weighted\ Average\ Recall = \frac{\sum_{i=1}^{m}|Y_{i}| \frac{tp_{i}}{tp_{i} + fn_{i}}}{\sum_{i}^{m}|Y_{i}|} 
\label{eq:Recall}
\end{align}

\begin{align}
Weighted\ Average\ F1-score = \frac{\sum_{i=1}^{m}|y_{i}| \frac{2tp_{i}}{2tp_{i} + fp_{i} + fn_{i}}}{\sum_{i}^{m}|y_{i}|}  
\label{eq:F1-score}
\end{align}

\section{Results and Discussion}
The performance of the proposed defect detection framework is investigated and reported from two aspects. The first set of results analyzes the performance of RCLBP with the different classifiers on the noise free NEU dataset and compares them with that of the LBP and CLBP frameworks. These results are presented in Table \ref{table:nonlin} in terms of the three evaluation metrics described in section \ref{table:evaluation metrics}. 

Comparing the results of LBP, CLBP and RCLBP in terms of the weighted average precision, recall and F1-score measures in Table \ref{table:nonlin} illustrates that RCLBP outperforms LBP and CLBP on the noise free NEU dataset. LBP when combined with the classifiers has a relatively moderate performance, as opposed to CLBP because the LBP features have a relatively lower discriminative power since only the sign component is encoded in the feature extraction. LBP thus is not able to clearly distinguish between inter-class defect similarities and intra-class defect differences. RCLBP, however with the classifiers has a relatively higher defect detection performance in terms of the evaluation metrics. This can be explained by the combination of the denoising technique which removes the natural noise already within the NEU defect dataset and the CLBP which extracts the discriminative defect features to clearly distinguish between the defect classes. All the classifiers KNN, SVM, DT, GNB and RF show comparable performance in building the defect detection models. The better performance of the proposed framework shows that the framework is clearly able to distinguish between the inter-class defect similarities and intra-class defect differences in the presence of the background interference. 

\begin{table}[h!]
\caption{Performance comparison of LBP, CLBP and RCLBP on noise-free NEU-DET dataset.} 
\centering 
\footnotesize
{\begin{tabular}{c c c c} 
\hline \hline 
\textbf{Classifier} &\textbf{LBP} &\textbf{CLBP} &\textbf{RCLBP}\\ [0.3ex]
 \hline \hline
\multicolumn{4}{c}{\text{Weighted Average Precision Comparison}} \\  \hline
KNN &0.89 &0.95 &0.98\\
SVM &0.87 &0.95 &0.98\\
DT &0.84 &0.88 &0.95\\
GNB &0.78 &0.95 &0.97\\
RF &0.89 &0.95 &0.97\\
\hline 
\hline
\multicolumn{4}{c}{\text{Weighted Average Recall Comparison}} \\  \hline 
KNN &0.88 &0.94 &0.98\\
SVM &0.87 &0.94 &0.98\\
DT &0.83 &0.88 &0.95\\
GNB &0.77 &0.95 &0.97\\
RF &0.88 &0.95 &0.97\\
\hline \hline \multicolumn{4}{c}{\text{Weighted Average F1-score Comparison}} \\  \hline 
KNN &0.88 &0.94 &0.98\\
SVM &0.87 &0.94 &0.98\\
DT &0.83 &0.88 &0.95\\
GNB &0.77 &0.95 &0.97\\
RF &0.88 &0.95 &0.97\\
\hline 
\end{tabular}}
\label{table:nonlin}
\end{table}

The second set of results analyzes the performance of LBP, CLBP and RCLBP on their robustness on the different SNR value of noise in terms of the three evaluation metrics. These results are presented in Table \ref{table:PPerrr3}.

At $SNR=50dB$, the weighted average precision, recall and F1-score of the LBP and CLBP drops comparably to the performance of the noise free experiment. The proposed approach RCLBP however maintains a comparable  performance to the performance on the noise free defect dataset. At $SNR=40dB$, RCLBP achieves comparable performance at  $SNR=50dB$. However, the performance of LBP drops considerably with the CLBP providing a relatively moderate performance. At $SNR=30dB$, the performance of the LBP and CLBP diminishes in terms of the three evaluation metrics. RCLBP however achieves considerable performance with a weighted average precision and recall of about $0.78$ and $0.72$ respectively with the weighted average F1-score of $0.70$.
At $SNR=20dB$, RCLBP still achieves considerable performance with a weighted average precision and recall of about $0.39$ and $0.39$ respectively with the weighted average F1-score of $0.29$ under the influence of the highest additive Gaussian noise influence. 
The robustness of the proposed framework to noise can be attributed to the denoising framework which which removes the signal while retaining the textures and edges characterizing the defect classes.

\begin{table*}[ht]
\caption{Performance comparison of LBP, CLBP and RCLBP on noisy data with different SNR value.} 
\centering 
\footnotesize
{\begin{tabular}{>{\bfseries}c*{10}{c}}

\multirow{3}{*}{\bfseries Classifier} & \multicolumn{3}{c}{\bfseries Weighted Average Precision} & \multicolumn{3}{c}{\bfseries Weighted Average Recall} & \multicolumn{3}{c}{\bfseries Weighted Average F1-score} \\ \cmidrule(lr){2-4} \cmidrule(lr){5-7} \cmidrule(lr){8-10}

& \textbf{LBP} & \textbf{CLBP} & \textbf{RCLBP} & \textbf{LBP} & \textbf{CLBP} & \textbf{RCLBP} & \textbf{LBP} & \textbf{CLBP} & \textbf{RCLBP} \\ \midrule
\multicolumn{10}{c}{\text{SNR=50}} \\  
 \hline
KNN &0.45 &0.90 &0.98 &0.58 &0.89 &0.98 &0.48 &0.88 &0.98\\
SVM &0.55 &0.91 &0.98 &0.54 &0.90 &0.98 &0.45 &0.89 &0.98\\
DT &0.37 &0.89 &0.92 &0.53 &0.87 &0.92 &0.42 &0.86 &0.92\\
GNB &0.43 &0.69 &0.96 &0.54 &0.76 &0.96 &0.45 &0.71 &0.96\\
RF &0.43 &0.90 &0.98 &0.59 &0.84 &0.97 &0.49 &0.80 &0.97\\ 

\hline 
\multicolumn{10}{c}{\text{SNR=40}} \\  \hline
KNN &0.39 &0.78 &0.92 &0.57 &0.79 &0.91 &0.46 &0.74 &0.91\\
SVM &0.37 &0.88 &0.98 &0.49 &0.81 &0.97 &0.40 &0.75 &0.97\\
DT &0.35 &0.83 &0.87 &0.49 &0.80 &0.87 &0.40 &0.77 &0.87\\
GNB &0.42 &0.66 &0.93 &0.53 &0.75 &0.93 &0.44 &0.69 &0.93\\
RF &0.42 &0.86 &0.94 &0.57 &0.81 &0.94 &0.48 &0.75 &0.94\\

\hline
\multicolumn{10}{c}{\text{SNR=30}} \\  
\hline 
KNN &0.39 &0.65 &0.84 &0.38 &0.55 &0.81 &0.28 &0.48 &0.81\\
SVM &0.39 &0.50 &0.78 &0.36 &0.48 &0.75 &0.26 &0.38 &0.74\\
DT &0.32 &0.56 &0.76 &0.31 &0.52 &0.67 &0.23 &0.47 &0.65\\
GNB &0.26 &0.56 &0.71 &0.36 &0.61 &0.65 &0.26 &0.56 &0.59\\
RF &0.37 &0.66 &0.82 &0.35 &0.62 &0.74 &0.26 &0.57 &0.72\\

\hline
\multicolumn{10}{c}{\text{SNR=20}} \\  
\hline 
KNN &0.03 &0.03 &0.41 &0.17 &0.17 &0.43 &0.05 &0.05 &0.33\\
SVM &0.03 &0.03 &0.40 &0.17 &0.17 &0.40 &0.05 &0.05 &0.28\\
DT &0.03 &0.19 &0.33 &0.17 &0.18 &0.34 &0.05 &0.07 &0.24\\
GNB &0.06 &0.23 &0.38 &0.14 &0.20 &0.37 &0.09 &0.10 &0.27\\
RF &0.03 &0.19 &0.42 &0.17 &0.17 &0.42 &0.09 &0.10 &0.27\\
\hline
\end{tabular}}
\label{table:PPerrr3}
\end{table*}

\section{Conclusion and Future Work}
In this paper, we presented the Robust Completed Local Binary Pattern (RCLBP) framework which is a combination of the NL-means filter with wavelet thresholding and CLBP which extracts the noise robust features to characterize the defect classes. The RCLBP framework not only detected surface defects under the influence of the feature variations of the intraclass and interclass changes, but also achieved a relatively moderate recognition accuracy in the toughest situations with additive Gaussian noise. Our future work will seek to investigate the use of surface defect dataset with a varying scale and a high number of defect classes to build more robust and generalized defect detection models. Further research will also seek to explore synthetic data augmentation to introduce more diversity and increase the size of the training samples.

\section{Acknowledgement}
This work is supported by the Lockheed Martin corporation under fund number 234363 and the Air Force Research Laboratory (AFRL) and Office of Secretary of Defense (OSD) under agreement number FA8750-15-2-0116. This work is also partially supported by the NASA University Leadership Initiative (ULI) under grant number 80NSSC20M0161. The authors would also like to thank the support from the $OUSD(R\&E)/RT\&L$ under Cooperative Agreement Number W911NF-20-2-0261.


\bibliographystyle{IEEEtran}
\bibliography{sample}

\end{document}